\documentclass[pdflatex,sn-basic]{sn-jnl}

\usepackage{graphicx}%
\usepackage{multirow}%
\usepackage{amsmath,amssymb,amsfonts}%
\usepackage{amsthm}%
\usepackage{mathrsfs}%
\usepackage[title]{appendix}%
\usepackage{xcolor}%
\usepackage{textcomp}%
\usepackage{manyfoot}%
\usepackage{booktabs}%
\usepackage{algorithm}%
\usepackage{pdflscape}
\usepackage{listings}%

\usepackage[british,UKenglish]{babel}
\usepackage{natbib}
\usepackage{hyperref} 
\usepackage{array}
\usepackage{bm}
\usepackage[noend]{algpseudocode}
\graphicspath{ {images/} }
\usepackage{xparse}

\usepackage{hyphenat}
\PassOptionsToPackage{hyphens}{url}\usepackage{hyperref}

\definecolor{codegreen}{rgb}{0,0.6,0}
\definecolor{codegray}{rgb}{0.5,0.5,0.5}
\definecolor{codepurple}{rgb}{0.58,0,0.82}
\definecolor{backcolour}{rgb}{0.97,0.97,0.97}

\lstdefinestyle{python_jay}{
    backgroundcolor=\color{backcolour},
    commentstyle=\color{codegreen},
    keywordstyle=\color{blue},
    numberstyle=\tiny\color{codegray},
    stringstyle=\color{codepurple},
    basicstyle=\ttfamily\footnotesize,
    breakatwhitespace=false,
    breaklines=true,
    captionpos=b,
    keepspaces=true,
    numbers=left,
    numbersep=5pt,
    showspaces=false,
    showstringspaces=false,
    showtabs=false,
    tabsize=2
}

\NewDocumentCommand{\rot}{O{45} O{1em} m}{\makebox[#2][l]{\rotatebox{#1}{#3}}}%

\newcolumntype{x}[1]{>{\centering\let\newline\\\arraybackslash\hspace{0pt}}p{#1}}

\theoremstyle{definition}

\begin{document}

\title[On time series clustering with $k$-means]{On time series clustering with $k$-means}

\author{Christopher Holder$^{1,2}$, Anthony Bagnall$^{1}$, Jason Lines$^{2}$}

\affil{$^{1}$School of Electronics and Computer Science, University of Southampton, Southampton, UK}
\affil{$^{2}$School of Computing Sciences, University of East Anglia, Norwich, UK}

\affil{c.l.holder@soton.ac.uk, a.j.bagnall@soton.ac.uk, j.lines@uea.ac.uk}

\maketitle

\abstract{
There is a long history of research into time series clustering using distance-based partitional clustering. Many of the most popular algorithms adapt $k$-means (also known as Lloyd's algorithm) to exploit time dependencies in the data by specifying a time series distance function. However, these algorithms are often presented with $k$-means configured in various ways, altering key parameters such as the initialisation strategy. This variability makes it difficult to compare studies because $k$-means is known to be highly sensitive to its configuration. To address this, we propose a standard Lloyd's-based model for TSCL that adopts an end-to-end approach, incorporating a specialised distance function not only in the assignment step but also in the initialisation and stopping criteria. By doing so, we create a unified structure for comparing seven popular Lloyd's-based TSCL algorithms. This common framework enables us to more easily attribute differences in clustering performance to the distance function itself, rather than variations in the $k$-means configuration.}

\keywords{Time series clustering, k-means, Lloyds algorithm, elastic distances}

\maketitle

\section{Introduction}
Time series clustering (TSCL) is a perennially popular research topic: there have been over 1500 papers per year for the last five years matching the terms ``Time Series Clustering" (according to Web of Science\footnote{\url{www.webofknowledge.com}}). Partitional clustering algorithms such as $k$-means are commonly employed and adapted for the unique nature of time series~\citep{holder24clustering}. $k$-means is an iterative algorithm that uses a local search algorithm to find the ``best" clusters as defined by an objective function. It characterises clusters with centroids that are prototypes or exemplars (we use the terms interchangeably) that represent the middle of each cluster. It involves four key stages: initialisation of the prototypes, assigning instances to clusters based on distance to the prototypes, recalculating the prototype based on new cluster membership (also called averaging), then deciding whether to stop the algorithm based on a convergence criteria. There are many variants of this basic algorithm. To avoid confusion, we call this core version of $k$-means Lloyd's algorithm, based on the author widely acknowledged as having first proposed it~\citep{lloyds82algo}.  Lloyd's algorithm can also be used with medoids (prototypes from the training data)~\citep{holder23kmedoids}, but our focus is on $k$-means based approaches. Hence, we will  use the terms $k$-means and Lloyd's algorithm interchangeably. Lloyd's algorithm forms the basis for a large proportion of TSCL research (e.g.~\cite{lafabregue22endtoenddeeplearning,paparrizos16kshapes,petitjean11dba,javed20benchmark,kobylin20kmeansreview,zhang19ussl,petitjean12satelliteimageclustering,paparrizos17kmshape,ma19representationintscl,alqahtani21deepclusteringreview,niennattrakul07tsclmultimedia,paparrizos23odyssey,holder24clustering}). 

Outside of TSCL, in traditional clustering there have been numerous modifications introduced to both the initialisation and stopping criteria stages, with many of these modifications considered essential for achieving meaningful results (see~\cite{ikotun23kmeansreview, calebi13initreview} for excellent reviews of $k$-means algorithms for tabular data). The baseline Lloyd's algorithm, such as that implemented in the widely used \texttt{scikit-learn} library (\cite{scikit-learn})\footnote{\url{https://scikit-learn.org/stable/modules/generated/sklearn.cluster.KMeans.html}}, has options for these variants alongside commonly accepted default procedures. Our aim is to translate and assess findings from TSCL, where there is no defined default version of Lloyd's algorithm that is consistently adopted, using a standardised algorithmic base. Some researchers incorporate specific modifications suggested in traditional clustering literature~\citep{holder24clustering}, while others either do not specify the clustering set up or use the unmodified, original Lloyd's algorithm~\citep{paparrizos17kmshape}. Some studies use different configurations of Lloyd's-based algorithms within the same paper (e.g.~\cite{javed20benchmark}). Our base hypothesis is that meaningful comparison of the impact of time series specific changes to Lloyd's-based algorithms may be confounded by variation in standard settings, and that experiments that do not employ the standard best practice for general clustering will lead to worse results that may obfuscate meaningful differences in time series specific algorithms. 

The majority of research into TSCL using partitional clustering has focused on either the distance function~\citep{holder24clustering} or the method for finding prototypes, often called the averaging method~\citep{petitjean11dba}. However, the distances and averaging method also impact on the initialisation and stopping criteria. The TSCL literature is often unclear on whether these steps have been adapted to the distance function adopted. 

Our primary contribution is to summarise variants of Lloyd's algorithm used in the TSCL literature, to assess the impact of variants of initialisation and stopping condition and to describe the default structure we adopt for all our subsequent research, which we hope other researchers will also adopt. Our version of Lloyd's algorithm uses a specified distance function in an end-to-end manner for all stages where similarity between series is required. We then compare some well known TSCL algorithms using the UCR archive of univariate time series problems~\citep{dau19ucr}\footnote{\url{https://timeseriesclassification.com}} under this scheme.  


The structure of this paper is as follows. In Section~\ref{sec:lloyds} we review the variants of the four stages of the general Lloyd's algorithm and review recent TSCL literature that propose variants of $k$-means Lloyd's clustering. In Section~\ref{sec:design} we describe alternative experimental designs and performance measures. 
Section~\ref{sec:lloyds-experiments} assesses the impact of common parameters for TSCL and proposes a default set up.
In Section~\ref{sec:results} we experimentally assess the variants earlier identified and arrive at a default. 

\section{Lloyd's $k$-means clustering}
\label{sec:lloyds}

\begin{figure}[h]
    \centering
    \includegraphics[width=1.0\linewidth]{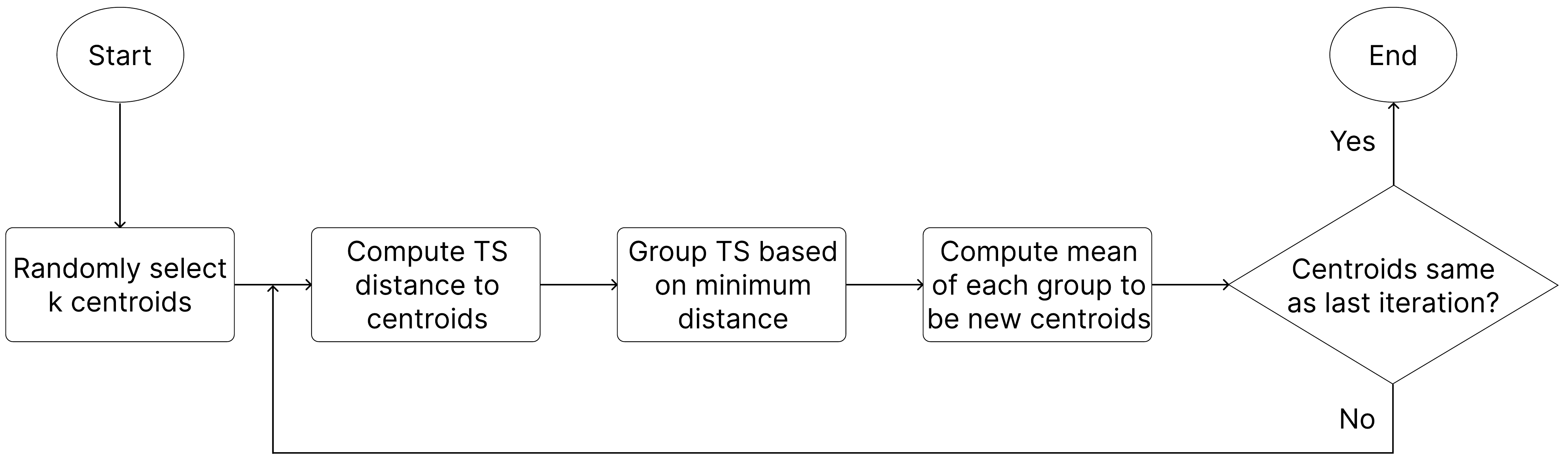}
    \caption{A flow diagram of the $k$-means algorithm.}
    \label{fig:k-means-flow-diagram}
\end{figure}

Given a dataset of $n$ time series $D = \{x_1, x_2, \ldots, x_{n}\}$ the goal of $k$-means is to form $k$ clusters such that the objective function is minimised. The objective function assumes the number of clusters, $k$, is defined and that we are trying to find $k$ exemplars, $C=\{c_1, c_2, \ldots, c_k\}$ for each cluster. The $k$-means objective function is then called the sum of squared error, SSE,

\begin{equation} SSE = \sum_{i=1}^{k} \sum_{j=1}^{n_i} d(x_{j}^{(i)}, c_i)^2.
\label{eq:sse}
\end{equation}
where $n_i$ is the number of time series assigned to cluster $i$, $x_{j}^{(i)}$ is the $jth$ time series assigned to $ith$ cluster, $c_i$ is the $ith$ prototype, and $d$ is a distance measure between two series. For standard $k$-means clustering, the  Euclidean distance is used for convergence reasons~\citep{bottou94convergence}. The SSE is sometimes rather confusingly referred to as inertia. The optimisation problem is to find the set of exemplars that minimise the objective. Lloyd's algorithm follows an expectation-maximisation like scheme of assigning series to the closest prototype, then recalculating the prototype, until convergence.  Figure~\ref{fig:k-means-flow-diagram} shows a flow diagram of the described $k$-means algorithm. 
The key steps are: 
\begin{enumerate}
    \item \textbf{initialisation}: set $k$ and create an initial prototype for each cluster.
    \item \textbf{assignment}: assign each instance to the nearest prototype, based on a distance function.
    \item \textbf{update}: recalculate the prototype based on the new assignment.
    \item \textbf{stop or repeat}: determine whether to continue with a further iteration.
\end{enumerate}
For tabular data, assignment for $k$-means uses the Euclidean distance for $d$. Time series specific distance functions is an active area of research in its own right, with new distance function variants still being proposed (e.g.~\cite{herrmann23adtw}. See~\cite{holder24clustering} for a survey and evaluation of distance based clustering.  
The update step involves averaging, and for standard $k$-means this means taking the arithmetic mean of the series assigned to a specific cluster. Again, algorithms for time series averaging is an active field of interest in its own right (e.g. ~\cite{petitjean11dba, holder23kmedoids,fawaz24weighted, fawaz23shapedba}) and is related to distances. Our focus is on initialisation and stopping criteria, and how they have been and should be used in TSCL.  

\subsubsection*{Initialisation}

The initialisation problem in $k$-means is twofold: defining the correct value of $k$, and defining the initial exemplars~\citep{ikotun23kmeansreview}. How to define the correct value of $k$ is an open question, but techniques such as the elbow method, Silhouette coefficient~\citep{kaufman90silhouettecoeff}, Canopy algorithm~\citep{esteves11canopyinit} and the Gap statistic algorithm~\citep{walther02gapinit} have been proposed to find a appropriate value of $k$. All the research we review assumes $k$ is known prior to clustering. Usually, $k$ is the number of classes in a classification problem.

The best method for selecting the initial exemplars is also an on going research question. However, several techniques have been proposed in the literature, including random selection, furthest point heuristic, sorting heuristic, density-based, projection-base and splitting techniques. These and other initialisation techniques are described in~\cite{calebi13initreview}, where a comparison was performed on 32 real and 12,228 synthetic datasets. The three methods most used in TSCL are:

\begin{itemize}
    \item \textit{Forgy} initialisation chooses $k$ random prototypes from the dataset to be the initial exemplars~\citep{forgy65forgyinit}.
    \item \textit{Random} initialisation selects $k$ initial prototypes by randomly assigning values within the data range, not necessarily corresponding to actual train instances. 
    \item \textit{$k$-means++} initialisation starts by choosing the first exemplar randomly from the training data. Each subsequent exemplar is then chosen based on a probability proportional to the distance from the nearest exemplar already selected~\citep{arthur07kmeansplusplus}. 
    \item \textit{Greedy $k$-means++} is a variant of $k$-means++ that, instead of using a probabilistic selection process for each new prototype, deterministically finds the time series in the dataset that maximises the minimum distance from the already chosen prototypes~\citep{arthur07kmeansplusplus}. 
\end{itemize}

Whilst~\cite{calebi13initreview} did not provide a definitive answer to the best initialisation strategy, some general conclusions were drawn. First, initialisation strategies such as Forgy and random initialisation should be avoided because single runs using such a random initialisation can easily lead to premature convergence at a poor local optima. Instead,~\cite{calebi13initreview} recommended approaches such as greedy $k$-means++. Alternatively, Forgy or random should be used with restarts: repeat the clustering process then choose the clustering with the lowest objective function value. Others have found similar results, such as~\cite{ahmed20initreview,franti199initandrepeats,ting04randominitrestart}, who also found that randomly generated prototypes lead to early convergence. 

\subsubsection*{Stopping conditions}

Lloyd's algorithm using squared Euclidean distance has been proven to always converge in a finite number of iterations~\citep{lloyds82algo} although this convergence may not necessarily lead to the global optimum. Lloyd's algorithm considers convergence to be achieved when either the cluster membership does not change or the change in objective function is very small between iterations. The objective function in this context is often referred to as the inertia, although it is in fact the SSE defined in Equation~\ref{eq:sse}. 
\subsubsection*{Empty clusters}

The performance of Lloyd's algorithm on any given dataset is dependent on the number of clusters specified~\citep{ikotun23kmeansreview}. One question rarely addressed is what happens if $k$ is fixed before hand, but during execution of the algorithm one or more clusters contains empty clusters. Once empty, a cluster prototype will not change, and will probably remain empty. This issue is particularly relevant when using random initialisation strategies, since it can skew the objective function. The traditional Lloyd's algorithm has no provision for dealing with empty clusters, since it is very unlikely to happen with Euclidean distance and a pathological initialisation. Accordingly, in many implementations, the presence of empty clusters is ignored. The general advice to practitioners is to use a better initialisation strategy. An alternative is, when an empty cluster forms, assign a random instance as a prototype during the run, or find one through, for example, finding the assignment that reduces SSE by the largest amount. The latter is the strategy that scikit-learn (\cite{scikit-learn}) employs to handle empty clusters in their $k$-means implementation.

\subsection{Time series Lloyd's clustering}
\label{sec:tscl}

We have identified seven variants of Lloyd's algorithm in the TSCL literature. 

\begin{itemize}
    \item \textit{$k$-means} is the traditional $k$-means algorithm that uses the squared Euclidean distance and the arithmetic mean, which minimises the sum of squared Euclidean distances (\cite{lloyds82algo}).
    \item \textit{$k$-means-DTW}: dynamic time warping (DTW) with $k$ means has been a popular baseline algorithm. The basic usage with Lloyd's is by replacing the Eucliden distance in the assignment phase, but keeping the arithmetic averaging stage (\cite{niennattrakul07tsclmultimedia}).
    \item \textit{$k$-means-DBA}: $k$-means with DTW barycentre averaging (\cite{petitjean11dba}) adapts both the assignment and the averaging stage to use and minimise DTW. 
    \item \textit{$k$-SC} is another variant of $k$-means that utilises spectral centroids in the distance calculation with an averaging technique that minimises the spectral centroid distance (\cite{jaewon11ksc}).
    \item \textit{$k$-shapes} is a variant that uses the shape-based distance (SBD) and employs a shape extraction algorithm to derive an average that minimises this distance (\cite{paparrizos16kshapes}).
    \item \textit{$k$-means-soft-dba} is a variant of $k$-means that uses the soft-DTW distance and minimises over the soft-DTW distance using the soft-DTW barycentre average (\cite{cuturi2017softdtw}).
    \item \textit{$k$-means-MSM} employs the Move-Split-Merge (MSM) distance (\cite{stefan13msm}), rather than DTW, and was found to be one of the most effective approaches in a comparative study (\cite{holder24clustering}). 
\end{itemize}


Table~\ref{tab:variants} highlights the specific configuration decisions made for each Lloyd's technique in a range of related papers. We observe that no two experiments use the same configuration of initialisation and stopping. We conducted an extensive search and found it is very rare to find articles to use the same Lloyd's configurations. We believe a significant factor contributing to this is the frequent reference to a ``default'' or ``traditional'' version of $k$-means (Lloyd's algorithm), without any clear authority defining this ``default'' version.

\begin{table}[h]
\centering
\makebox[\textwidth][c]{%
\begin{tabular}{|p{2cm}|p{1cm}|p{1.5cm}|p{3cm}|p{3cm}|p{1cm}|p{2cm}|}
\hline
\textbf{Reference/ Algorithm} & \textbf{Num. Cites} & \textbf{Init} & \textbf{Distance} & \textbf{Averaging} & \textbf{Max iters} & \textbf{Early stopping} \\ \hline
\cite{niennattrakul07tsclmultimedia} \textbf{$k$-means-DTW} & 263 & Random & Euclidean, DTW & Mean & - & No change in clusters \\ \hline
\cite{petitjean11dba} \hspace{0.4cm}\textbf{$k$-means-DBA}& 1248 & Forgy & DTW & DBA & 10 & - \\ \hline
\cite{jaewon11ksc} \hspace{0.4cm}\textbf{$k$-SC}& 1329  & Random & SC & SC & - & No change \\ \hline
\cite{paparrizos16kshapes} \textbf{$k$-shapes}& 504 & Random & Euclidean, SBD, DTW, KSC dist & Mean, DBA, Shape extraction, KSC average & 100 & No change  \\ \hline
\cite{paparrizos17kmshape} & 144 & Random & Euclidean, SBD, DTW, KSC dist & Mean, DBA, Shape extraction, KSC average & 100 & No change  \\ \hline
\cite{cuturi2017softdtw} \hspace{0.4cm}\textbf{soft-DBA}& 766  & - & soft-DTW & soft-DBA & 30 & -\\ \hline

\cite{zhang19ussl} & 95 & - & Euclidean, SBD, DTW, KSC dist & Mean, DBA, Shape extraction, KSC average & - & - \\ \hline
\cite{ma19representationintscl} & 213 & Random with 5 restarts & Euclidean, SBD, DTW, KSC dist & Mean, DBA, Shape extraction, KSC average & - & - \\ \hline
\cite{kobylin20kmeansreview} & 2 & Forgy & DTW, Euclidean, soft-DTW & Mean, DBA, soft-DBA & - & - \\ \hline
\cite{javed20benchmark} & 148 & Forgy & DTW, Euclidean, SBD & Mean, DBA, Shape extraction & 15 & No change \\ \hline
\cite{alqahtani21deepclusteringreview} & 50 & Forgy & Euclidean, SBD, DTW, KSC dist & Euclidean, SBD, DTW, KSC dist & - & - \\ \hline
\cite{lafabregue22endtoenddeeplearning} & 50 & $k$-means++ & Euclidean, SBD, DTW & Mean, DBA, Shape extraction & 200 & Inertia \\ \hline
\cite{paparrizos23odyssey} & 1 & Random & Euclidean, SBD, DTW, KSC dist & Mean, DBA, Shape extraction, KSC average & 300 & No change  \\ \hline
\cite{holder24clustering} \textbf{$k$-means-msm} & 27 & Forgy with 10 restarts &  MSM+ others & Mean & 300 &  Inertia \\ \hline
\end{tabular}
}
\caption{A sample of Lloyd's clustering algorithm configuration for TSCL. If a cell contains a ``—'', the corresponding value is unspecified or not included the paper or related material.
}
\label{tab:variants}
\end{table}




\section{Experimental design}
\label{sec:design}

\begin{figure}
    \centering
        \includegraphics[width=0.7\textwidth]{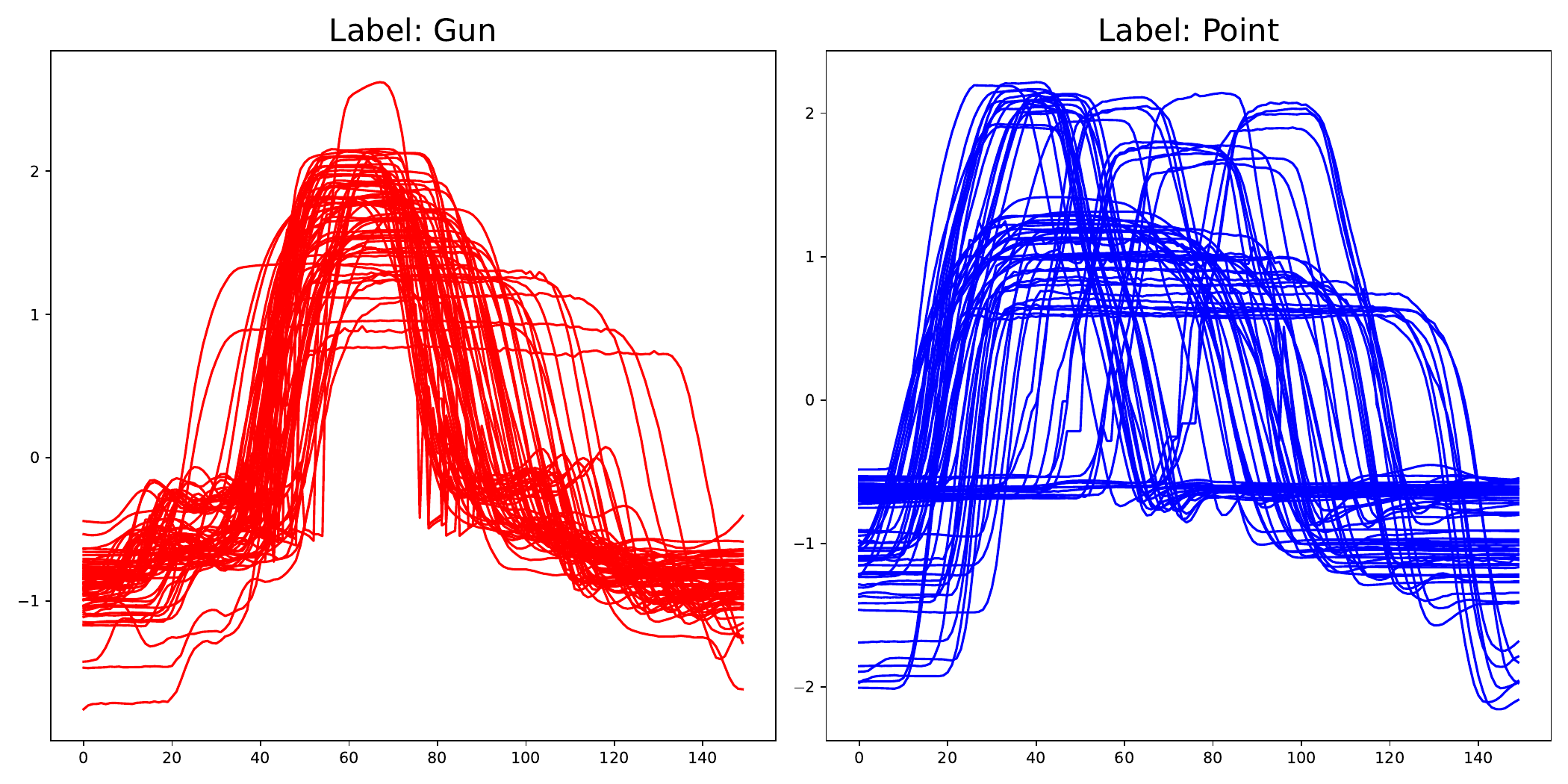}\\
(a)    \\
        \includegraphics[width=0.7\textwidth]{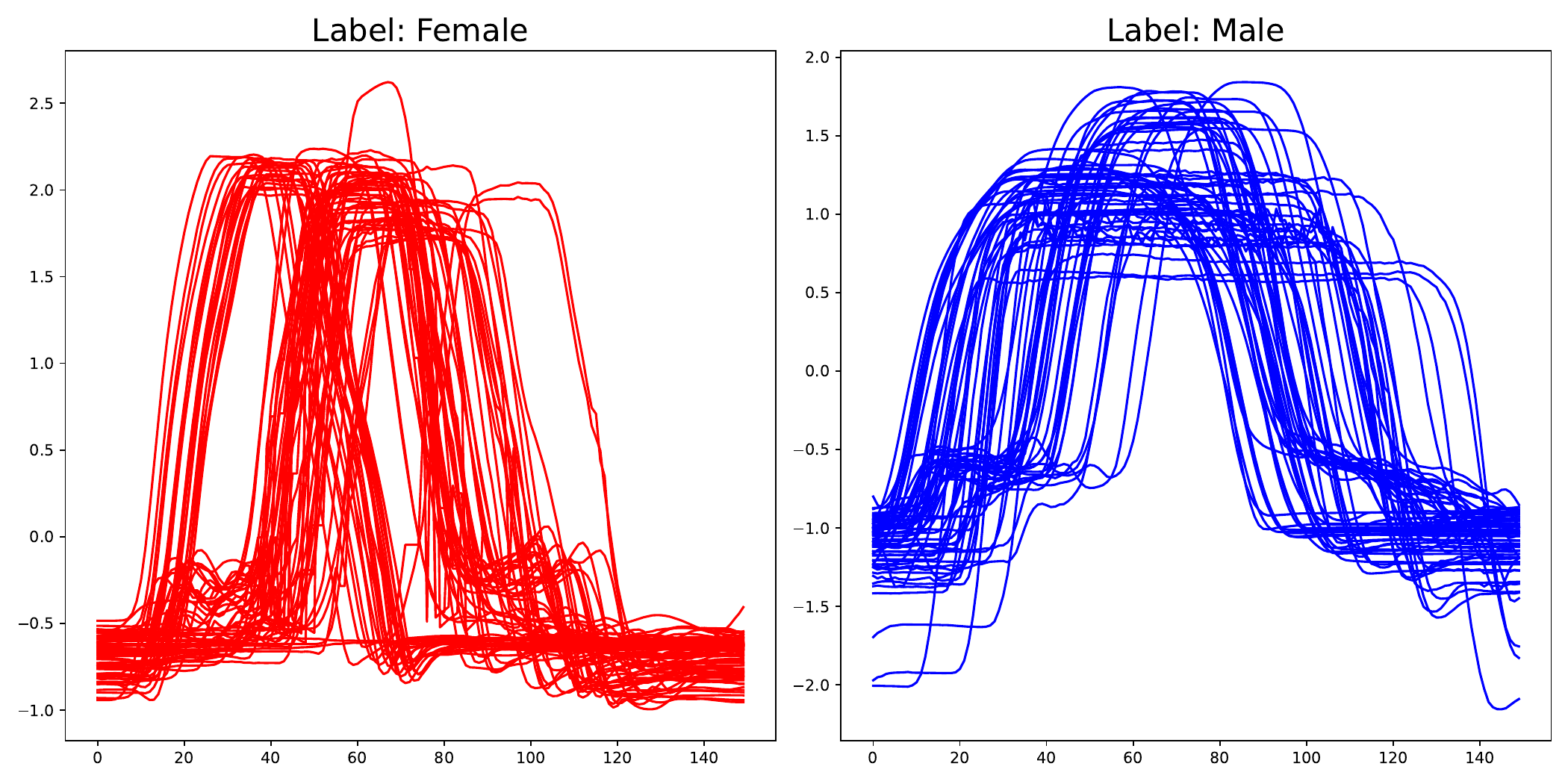}\\
        (b)\\
    \caption{Examples of different clusterings of the GunPoint dataset. The top clusters (a) are grouped by class label while the bottom clusters (b) are grouped by the participant.}
    \label{fig:gunpoint-clusters}
\end{figure}

To assess clustering algorithms, we need to first specify how to compare clusters. 
Evaluating clusters is inherently challenging, as ``clusters are, in large part, in the eye of the beholder'' (\cite{estivill02somanyclusteralgos}). This subjectivity means there is no single ``correct'' way to cluster a dataset, making the evaluation of clustering methods particularly complex. Consider, for example, the  GunPoint time series dataset from the UCR archive (\cite{dau19ucr}). The GunPoint dataset was originally created to capture the motion of actors performing two distinct actions: pointing a gun or pointing with their fingers. This dataset was generated by recording two actors (one male and one female) executing these actions over a five-second period, resulting in a sequence of 150 frames per action. The x-axis coordinate of the hand was extracted from each frame to form the time series data. The dataset was labeled with two class labels: ``gun'' and ``point'' reflecting the actor's action. 
Figure~\ref{fig:gunpoint-clusters} shows two different clustering of the GunPoint dataset. The first 

In a time series classification task, the objective is to separate the time series into these two categories. However, in the context of clustering, where the labels are unknown, this dataset can be interpreted in multiple valid ways. Figure~\ref{fig:gunpoint-clusters} illustrates two different interpretations of the GunPoint dataset. The top figures show the series grouped by class label, the bottom two by the sex of the participant. Both could be argued as valid clustering. However, if we use the class labels to assess a clustering, then the top groupings will be 100\% accurate, the bottom two 50\%. 

This is one example of the complexity of cluster evaluation using classification data. There are other datasets in the UCR archive that are equally likely to confound the evaluation. However, for consistency with the related work, we continue to use all the equal length UCR datasets in our evaluation. 

Given a ground truth label, there are a range of ways of assessing how well a clustering matches the ground truth.

\textbf{Clustering accuracy (CL-ACC)} is the number of correct predictions divided by the total number of cases. To determine whether a cluster prediction is correct, each cluster has to be assigned to its best matching class value. This can be done naively, taking the maximum accuracy from every permutation of cluster and class value assignment $S_k$.
$$
    CL-ACC(y,\hat{y}) = \max_{s \in S_k} \frac{1}{|y|} \sum_{i=1}^{|y|}
    \begin{cases}
        1, & y_i = s(\hat{y}_i) \\
        0, & \text{otherwise}
    \end{cases}
    \label{eqn:accuracy}
$$

Checking every permutation like this is prohibitively expensive, however, and can be done more efficiently using combinatorial optimization algorithm for the assignment problem. A contingency matrix of cluster assignments and class values is created, and turned into a cost matrix by subtracting each value of the contingency matrix from the maximum value. Clusters are then assigned using an algorithm such as the Hungarian algorithm on the cost matrix. If the class value of a case matches the assigned class value of its cluster, the prediction is deemed correct, else it is incorrect. As classes can only have one assigned cluster each, all cases in unassigned clusters due to a difference in a number of clusters and class values are counted as incorrect.

The \textbf{rand index (RI)} works by measuring the similarity between two sets of labels. This could be between the labels produced by different clustering models (thereby allowing direct comparison) or between the ground truth labels and those the model produced. The rand index is the number of pairs that agree on a label divided by the total number of pairs. 
The rand index is popular and simple. However, it is not comparable across problems with different number of clusters. The \textbf{adjusted rand index (ARI)} compensates for this by adjusting the RI based on the expected scores on a purely random model.

The \textbf{mutual information (MI)}, is a function that measures the agreement of a clustering and a true labelling, based on entropy. 
\textbf{Normalised mutual information (NMI)} rescales MI onto $[0,1]$, and \textbf{adjusted mutual information (AMI)} adjusts the MI to account for the class distribution.

The UCR archive contains 112 datasets. We fit clusterers to each data and assess performance with these four measures. 

There are two further design decisions: whether to normalise all, and whether to merge the train and test data or train and test on separate data samples.  The issue of whether to always normalise is an open question in time series machine learning, since discriminatory features may be in the scale or variance, but some previous work has advocated that normalisation should always occur. For example, a 2012 paper that has been cited over 1000 times states that {\em ``In order to make meaningful comparisons between two time series, both must be normalized"}~\citep{rakthanmanon13trillionsubsequence}. Our choice is always to normalise, since our primary interest is to comparing different approaches to TSCL using a consistent methodology, rather than trying to maximise performance for a single use case. 

The UCR time series data are provided with default training and testing splits as evaluating on unseen and consistent test data is essential for any classification comparison. The issue is less clear cut with clustering, which is used more as an exploratory tool than a predictive model. Many comparative studies (e.g.~\cite{javed20benchmark}) combine train and test data to perform evaluation on a single data set, while other recent research has used training data only to find clusters and then used test data for evaluation (e.g.~\cite{lafabregue22endtoenddeeplearning}). We prefer the latter configuration, because clustering is often used as part of a wider pipeline in tasks such as segmentation, and in these  cases it is important to be able to assess performance on unseen data. 

All experiments are conducted with the \texttt{aeon} toolkit~\citep{middlehurst2024aeonpythontoolkitlearning}. A notebook associated with the paper is available using the $TimeSeriesKMeans$ clustering algorithm, which can be configured to replicate all the $k$-means algorithms described in this paper. A notebook demonstrating the running of these experiments is in the associated notebook\footnote{\url{https://github.com/time-series-machine-learning/tsml-eval/tree/main/tsml_eval/publications}}.

To compare multiple clusterers on multiple datasets, we use the rank ordering of the algorithms we use an adaptation of the critical difference (CD) diagram~\citep{demsar06comparisons}, replacing the post-hoc Nemenyi test with a comparison of all classifiers using pairwise Wilcoxon signed-rank tests, and cliques formed using the Holm correction recommended by~\cite{garcia08pairwise} and~\cite{benavoli16pairwise}. We use $\alpha=0.05$ for all hypothesis tests. Critical difference diagrams such as those shown in Figure~\ref{fig:cd-init-algorithms} display the algorithms ordered by average rank of the statistic in question and the groups of algorithms between which there is no significant difference (cliques).

\section{A TSCL configuration for Lloyd's-based algorithms}
\label{sec:lloyds-experiments}

As we are unable to find a clear definition of a default variant of Lloyd's algorithm for TSCL, we take it upon ourselves to recommend an explicit default implementation of Lloyd's algorithm for TSCL. We do this because of the known sensitivity of $k$-means to its parameters. By defining a default setting we will eliminate differences in the performance of the Lloyd's algorithm caused by the clustering configuration, thus allowing us to more clearly compare the time series component of the algorithms. Our goal is not to create the best version of Lloyd's algorithm, as the literature suggests there is no such optimal configuration (\cite{calebi13initreview}), but rather to develop a version that is both consistent and robust. Our configuration choices are motivated by traditional clustering literature and experimental evidence using the UCR archive. 

\subsection{Initialisation Strategy}

We consider the following initialisation strategies: Greedy $k$-means++; random; Forgy; random restarts and Forgy restarts. 

To assess the initialisation techniques, we conducted a series of simple experiments using the $k$-means with Euclidean distance with different initialisation techniques. We choose 10 restarts as recommended in~\cite{ting04randominitrestart}. 
\begin{figure}[htb]
    \centering
            \begin{tabular}{c c}
    \includegraphics[width=0.5\linewidth]{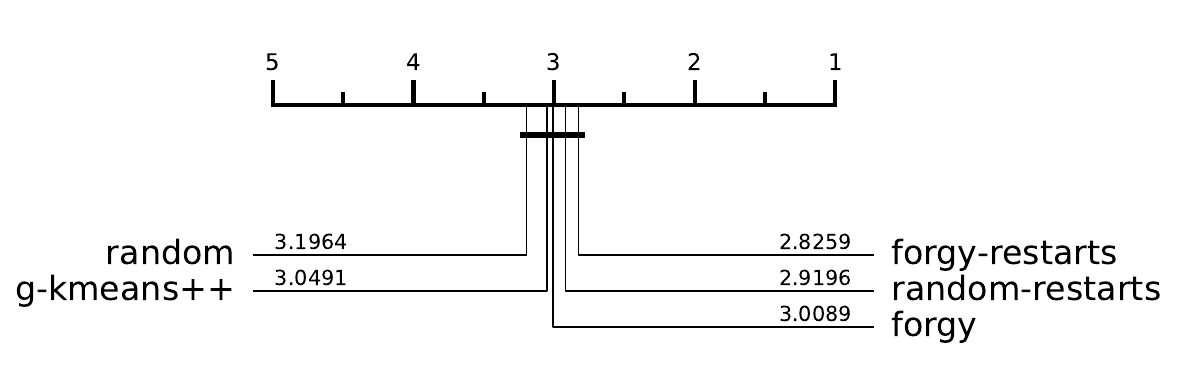} &
    \includegraphics[width=0.5\linewidth]{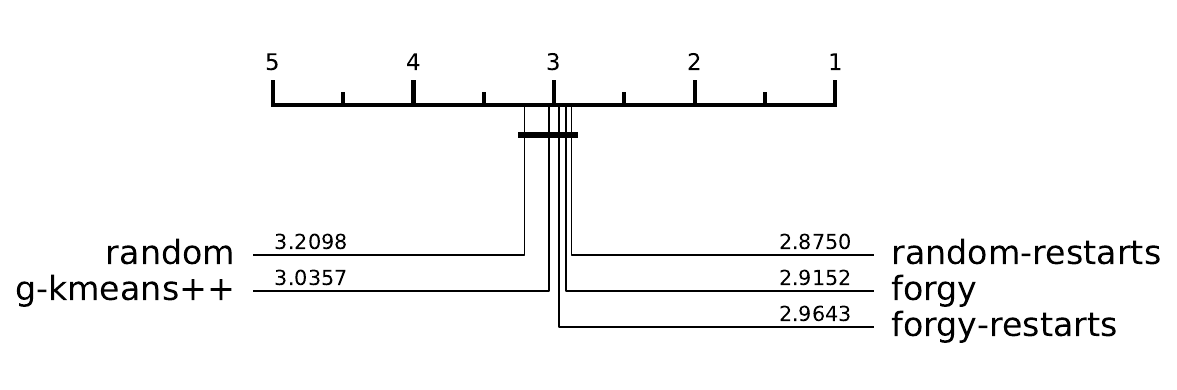}\\
    (a) Accuracy & (b) ARI \\
    \includegraphics[width=0.5\linewidth]{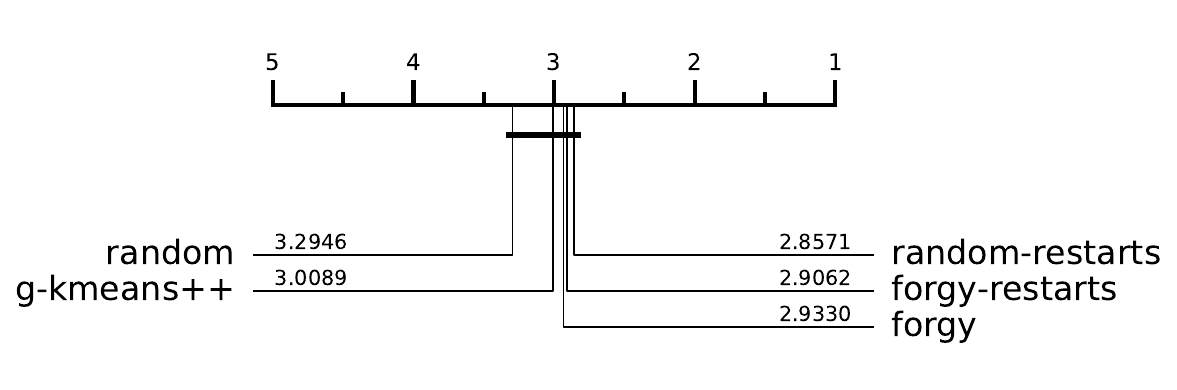} & 
    \includegraphics[width=0.5\linewidth]{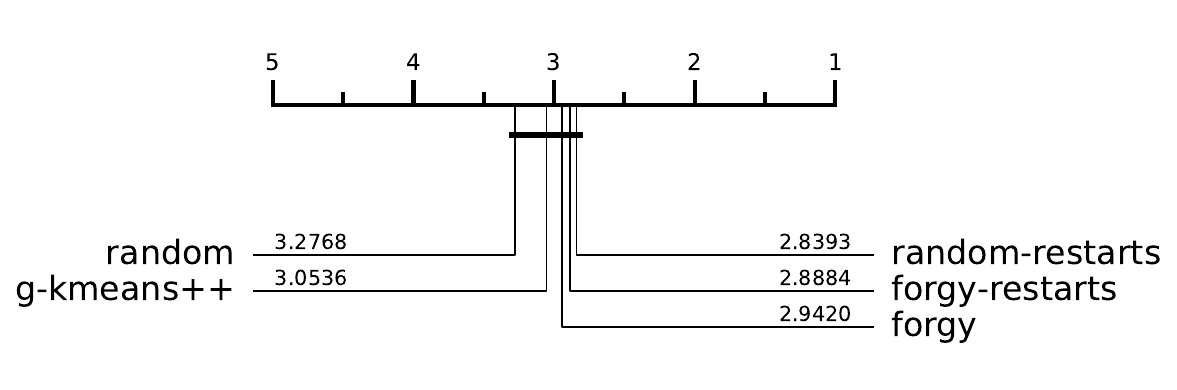} \\
    (c) AMI & (d) NMI \\
    
    \end{tabular}
    \caption{CD diagrams of different initialisation strategies for $k$-means over 112 datasets from the UCR archive using the combined test-train split. ``random'' refers to random initialisation, ``random-restarts'' refers to random initialisation with 10 restarts, where the restart with the lowest inertia is selected. ``forgy'' denotes Forgy initialisation, ``forgy-restarts'' represents Forgy initialisation with 10 restarts, and ``g-kmeans++'' denotes greedy $k$-means++.}
    \label{fig:cd-init-algorithms}
\end{figure}

\begin{table}[h]
    \vspace{0.25cm}
    \centering
    \begin{tabular}{|l|l|l|l|l|l|}
        \hline
         & ARI & AMI & CLACC & NMI & RI \\
        \hline
        random-restarts & \textbf{0.209} & \textbf{0.258} & 0.525 & \textbf{0.283} & \textbf{0.700} \\
        forgy-restarts & 0.208 & 0.257 & \textbf{0.526} & 0.283 & 0.699 \\
        random & 0.201 & 0.251 & 0.518 & 0.276 & 0.697 \\
        forgy & 0.203 & 0.255 & 0.521 & 0.281 & 0.696 \\
        g-$k$-means++ & 0.203 & 0.250 & 0.518 & 0.276 & 0.693 \\
        \hline
    \end{tabular}
    \vspace{0.25cm}
    \caption{Summary of initialisation strategies' average scores across multiple evaluation metrics over 112 datasets from the UCR archive using the combined test-train split.}
    \label{tab:init-algo-summary-mean-performance}
\end{table}

The CD diagrams in Figure~\ref{fig:cd-init-algorithms} show that for $k$-means, all the initialisation strategies fall into the same clique and therefore there is no significant difference between them. Table~\ref{tab:init-algo-summary-mean-performance} presents the specific average scores across the UCR archive. While the values in Table~\ref{tab:init-algo-summary-mean-performance} appear fairly similar across all metrics, assessing the variation of each measure guides our choice of recommended initialisation.

\begin{figure}[h]
    \centering
    \includegraphics[width=0.8\linewidth]{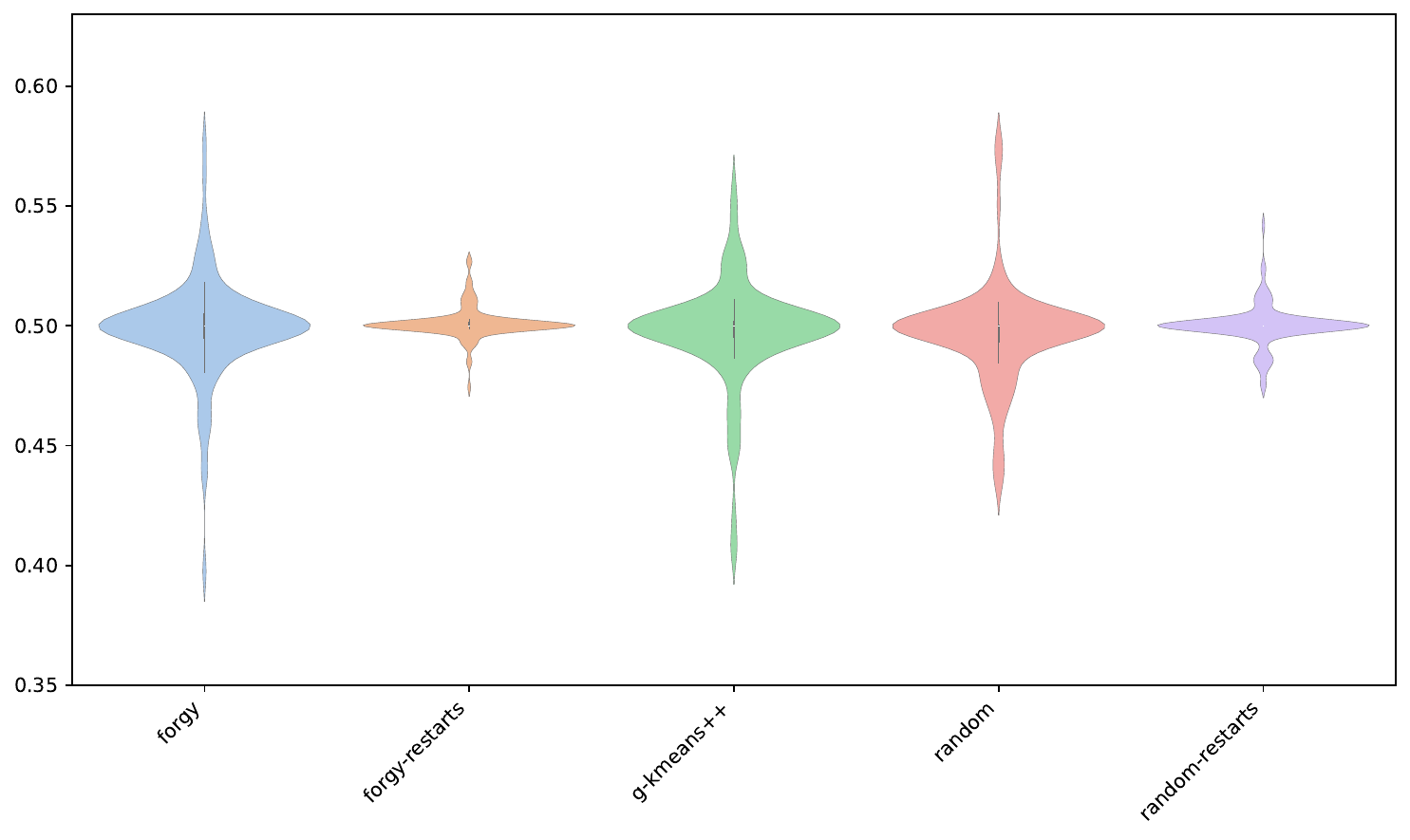}
    \caption{CLACC violin plot for different initialisation strategies over the 112 of the UCR archive using the combined test-train split.}
    \label{fig:clacc-init-algo-violin}
\end{figure}

Figure~\ref{fig:clacc-init-algo-violin} shows a violin plot of the different initialisation strategies across the UCR archive for accuracy. The data is the deviation from the median accuracies for each of the strategy. In Figure~\ref{fig:clacc-init-algo-violin}, Forgy, random, and greedy $k$-means++ display large variability in their performance, as demonstrated by the distribution of their plots. However, Forgy with 10 restarts and random with 10 restarts produce significantly more consistent results with very little variability across the UCR archive. This is the type of performance we seek for our experimental methodology, since we want to control for variability caused by the initialisation strategy.  

Based on Figure~\ref{fig:clacc-init-algo-violin}, Table~\ref{tab:init-algo-summary-mean-performance}, and Figure~\ref{fig:cd-init-algorithms}, we could choose either random restart or Forgy restart. We elect to use Forgy as it is the simplest option and because it is based on instances in the train data. This avoids having to specify methodology for resolving issues such as unequal length time series. 

\begin{figure}[h]
    \centering
    \includegraphics[width=0.8\linewidth]{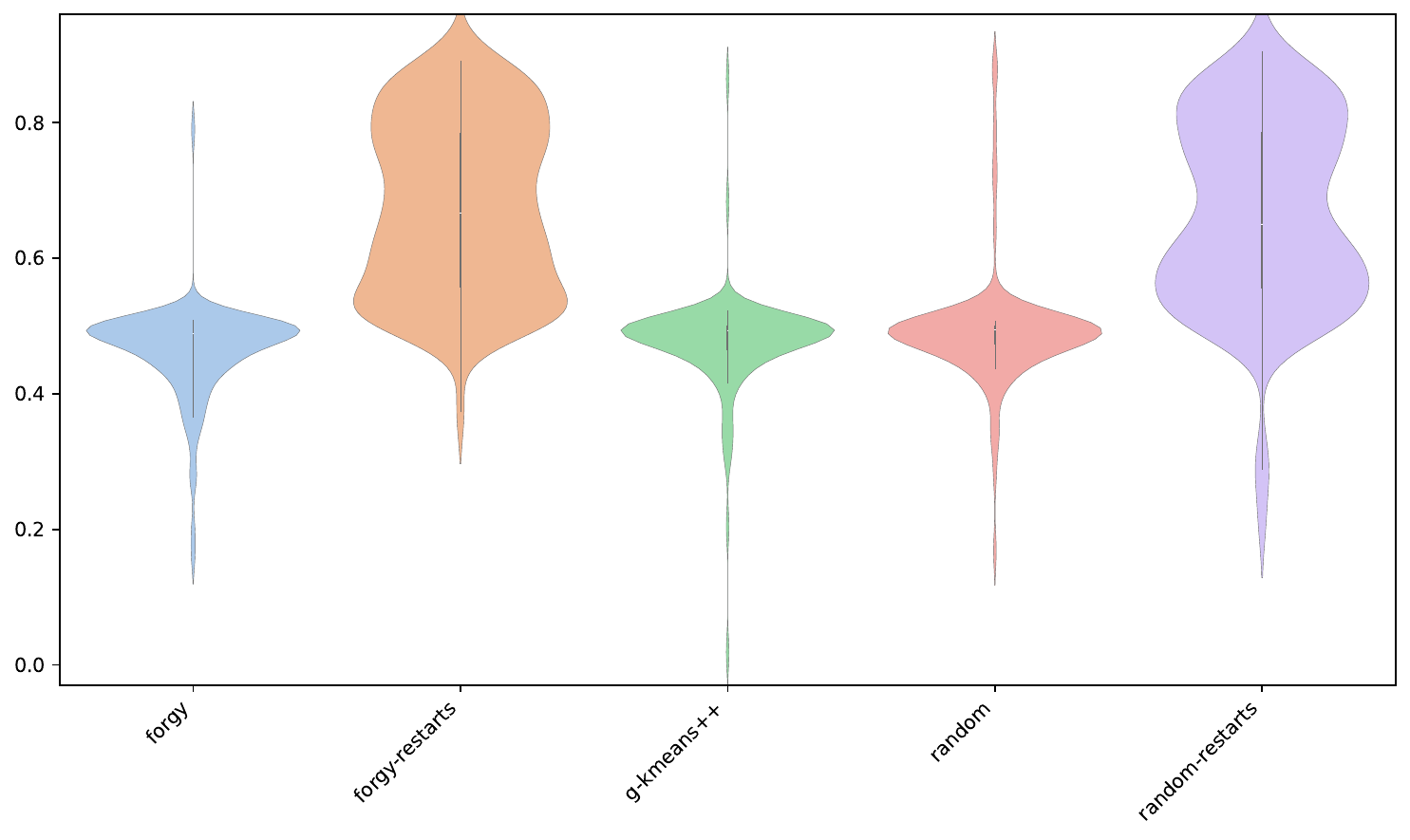}
    \caption{A violin plot to demonstrate the differences in run time for various initialisation strategies over the 112 UCR datasets with test-train split.}
    \label{fig:init-algo-fit-time}
\end{figure}

One of the disadvantages of using an initialisation with restarts is the increasing run time required to fit the model by the number of restarts. This is illustrated in Figure~\ref{fig:init-algo-fit-time}, where the distribution of run times for initialisation techniques that use restarts is significantly higher than for those that only run once. Nevertheless, we still believe it worthwhile to improve the consistency of results.

\subsection{Early Stopping Conditions}

The default Lloyd's algorithm is said to have converged when the Sum of Squared Errors (SSE) for a specific distance function does not change between iterations. If the SSE does not change, it indicates that the cluster assignments and, consequently, the centroids have not changed, signifying that the algorithm has converged. 

However, there are many alternative in addition to this stoppage condition, numerous early stopping conditions have been proposed for Lloyd's algorithm. The purpose of these early stopping conditions is to terminate the algorithm before full convergence is reached, thereby saving computation time. Early stopping is often utilised because Lloyd's algorithm exhibits diminishing returns with each iteration. Theoretically, as Lloyd's algorithm progresses towards convergence, the number of changes to cluster assignments should decrease with each iteration. This suggests that while a substantial number of changes might occur in the initial iterations, these changes should significantly diminish as the algorithm continues.

As a result, Lloyd's algorithm may reach a point where it updates very little between iterations, yet it takes a considerable amount of time to reach a final converged solution. There is also the possibility of it getting trapped oscillating between two local minima. Therefore, early stopping conditions aim to strike a balance between obtaining good, near-converged results while terminating at an appropriate time to save computational resources. It is common to employ a parameter for the maximum number of iterations. This stopping condition acts as a safety net to prevent the algorithm from running for an impractically long time. This parameter varies considerably between algorithms (See Table~\ref{tab:variants}) and could influence the performance of the clusterer.

A good default value for the maximum number of iterations is difficult to estimate because convergence depends on several factors: the number of clusters, the size of the dataset, and the initial prototypes selected. Additionally, some algorithms in the context of TSCL require more iterations than traditional Lloyd's due to the use of approximation strategies. For example, the $k$-means-DBA averaging stage employs DBA, which is an approximation of the average under DTW. The use of approximations can result in slower convergence.

With these factors considered, we aim to set our maximum iterations as high as possible while being conscious of our computational resources. Table~\ref{tab:mean-max-min-iterations} shows the average, minimum, and maximum iterations for computing squared Euclidean $k$-means over 112 datasets from the UCR archive using the combined test-train split. On average, a dataset in the UCR archive will converge in under 20 iterations. However, there are $9$ datasets that take, on average, more than $40$ iterations in their ``best iteration'' to converge. These datasets are shown in Table~\ref{tab:number-iterations-datasets}. Of the $9$ datasets, only $2$ exceed $50$ iterations on an average iteration.

\begin{table}[h]
\vspace{0.25cm}
\centering
    \begin{tabular}{|l|r|r|}
    \hline
     & Average Iterations & Best Iteration \\
    \hline
    Mean & 17.42 & 18.44 \\
    Min & 3.80 & 3.00 \\
    Max & 74.90 & 140.00 \\
    \hline
    \end{tabular}
\vspace{0.25cm}
\caption{Number of iterations required for the squared Euclidean distance clustering algorithm to converge without an early stopping condition on 112 datasets from the UCR archive using the combined test-train split.}
\label{tab:mean-max-min-iterations}
\end{table}

\begin{table}[h]
\vspace{0.25cm}
\centering
    \begin{tabular}{|l|r|r|}
    \hline
    Dataset & Best Iteration & Average Iterations \\
    \hline
    ElectricDevices & \textbf{140} & \textbf{74.9} \\
    Crop & 59 & 66.2 \\
    FaceAll & 52 & 37.0 \\
    UWaveGestureLibraryAll & 51 & 26.6 \\
    UWaveGestureLibraryZ & 51 & 46.0 \\
    FordA & 47 & 48.0 \\
    SemgHandSubjectCh2 & 47 & 31.8 \\
    UWaveGestureLibraryY & 46 & 43.3 \\
    FacesUCR & 42 & 37.4 \\
    \hline
    \end{tabular}
\vspace{0.25cm}
\caption{The $9$ datasets that averaged over $40$ iterations in their ``Best Iterations'' for the squared Euclidean $k$-means, out of 112 datasets from the UCR archive using the combined test-train split.}
\label{tab:number-iterations-datasets}
\end{table}
With these baseline statistics on the number of iterations it takes for the squared Euclidean distance to converge over the UCR archive, we elect to set our maximum number of iterations to $50$. While this number means that $5$ datasets would not have converged (as shown in Table~\ref{tab:number-iterations-datasets}), we are limited by computational resources. If we had unlimited computational resources, we would set the maximum number of iterations to $300$ or more to ensure convergence in every scenario. However, when considering the computational cost of experiments with such high iterations, which run with $10$ restarts, this value is not viable. It is also common to have a threshold for change in SSE/inertia, in case small changes caused by numerical calculations. We set our value of tolerance threshold to be $1 \times 10^{-6}$. This is the default for the \texttt{tslearn} package, although \texttt{scikit-learn} uses a value of $1 \times 10^{-4}$.

\subsection{Empty Clusters}
The performance of Lloyd's algorithm on any given dataset is dependent on the number of clusters specified~\citep{ikotun23kmeansreview}. In common with most of the literature, we assume the number of clusters in known beforehand in these experiments, and, if used in practice, would be set by a form of elbow finding. It is not uncommon for $k$-means to form empty clusters during the fit process. This is problematic because an empty cluster will never change in Lloyd's algorithm, and essentially reduces the number of clusters found by one. 

Whilst an empty cluster will only happen infrequently at initialisation with Euclidean distance and randomised centroids, elastic distances with alternative averaging algorithms quite frequently end in this situation. This can bias results and complicate evaluation, since all our metrics to compare algorithms assume an equal number of clusters.

We found no explicit explanation of how to deal with empty clusters in the TSCL literature.  The only acknowledgement of empty clusters forming in TSCL that we could find was in the source code provided by~\cite{javed20benchmark} where the formation of an empty cluster was used as a form of early stopping condition (although this is not mentioned in the paper). We do not believe this is a good criterion to express early convergence when using forms of random initialisation. If an empty cluster is formed early in the fit procedure then the cluster quality may be worse than a longer run.

In traditional clustering, there is limited literature on defining protocols for handling empty clusters. The general advice to practitioners is either to use a better initialisation strategy (e.g., $k$-means++) or to reduce the number of clusters~\citep{scikit-learn}. This does not fit our experimental set up. Hence, we adopt a strategy where, when an empty cluster is formed, we choose a time series from the dataset to become a new centroid. The choice of which time series to select is an important consideration. There are two common methods for selecting this new centroid: randomly choosing a series from the dataset or choosing a series that reduces inertia by the largest amount. The latter approach is that adopted by scikit-learn~\citep{scikit-learn} to handle empty clusters in their $k$-means implementation.

Randomly selecting a time series from the dataset to be a new centroid is a simple solution but has certain implications. Firstly, the selected value could be located in a similar position to the previous empty cluster, potentially leading to another empty cluster forming shortly thereafter, necessitating yet another random centroid selection. Secondly, at the point where the empty cluster forms, the algorithm is likely already partially converged towards a local optimum. Selecting a new random centroid could entirely change the optimum the algorithm was converging towards. To some extent, this could be considered equivalent to a complete restart of the algorithm with new initial centroids, leading to convergence towards a different optimum. As such, random selection may unintentionally bias clustering results. Therefore, we choose not to use this strategy.

Choosing a time series that reduces inertia by the largest amount as the new centroid is generally a more effective strategy than random selection because it directly targets the objective of the clustering algorithm. 

By selecting a time series that minimises inertia, the algorithm ensures that the new centroid contributes to a more optimal clustering configuration. This approach helps the algorithm maintain its progress toward convergence rather than potentially disrupting it with a random selection that could lead to a less efficient clustering outcome. Furthermore, it reduces the likelihood of forming another empty cluster, as the selected time series is likely to be situated in a region of the dataset where its inclusion will meaningfully improve cluster quality.

To identify the time series that would reduce the inertia by the largest amount, the time series that is furthest from its assigned cluster centroid should be chosen. This approach ensures that the new centroid is positioned in a way that maximally improves the overall clustering by reducing the distance of an outlier data point, thereby contributing the most to the reduction of inertia.

\subsection{Lloyd's Baseline}
\label{sec:lloyds-baseline-config}

Our final experimental design for Lloyd's algorithm uses Forgy with 10 restarts for initialisation, a tolerance for inertia for stopping and a reassignment of centroid for an empty cluster that minimises overall inertia. All three of these stages require a distance function to measure SSE/inertia. A large proportion of TSCL research examines the effect of different distance functions on the clusterer. Our implementation of $k$-means uses the same distance function throughout the process: we say it employs an end-to-end distance function.

\section{Comparison of Lloyd's-based TSCL algorithms}
\label{sec:results}

Each model shares Lloyd's-specific parameters. These parameters are detailed in Table~\ref{tab:lloyds-base-model-configuration}.

\begin{table}[h]
    \vspace{0.25cm}
    \centering
    \makebox[\textwidth][c]{%
    \begin{tabular}{|l|l|l|}
        \hline
         Algorithm & Distance & Centroid Computation\\
        \hline
        $k$-means-Euclidean & Euclidean & Arithmetic mean \\
        $k$-means-DTW & DTW & Arithmetic mean \\
        $k$-means-MSM & MSM & Arithmetic mean \\
        $k$-shape &  SBD & Shape extraction \\
        $k$-means-dba  & DTW & DBA \\
        $k$-SC & $k$-SC distance & $k$-SC average \\
        $k$-means-soft-dba & soft-DTW & soft-DBA \\
        \hline
    \end{tabular}%
    }
    \vspace{0.25cm}
    \caption{Baseline Lloyd's-based models parameters}
    \label{tab:lloyds-base-model-configuration}
\end{table}

Five of Lloyd's-specific parameters are kept constant (control variables). The independent variable for this baseline experiment is, therefore, the distance and averaging technique used.

In addition to Lloyd's-specific parameters, some models require additional parameters for their distance functions. These distance-specific parameters are summarised in Table~\ref{tab:lloyds-base-model-distance-configuration}. 

The $k$-SC distance measure includes a parameter, $max\_shift$, which is an integer ranging from $0$ to $m$, where $m$ is the length of the time series. This parameter controls the shifts that $k$-SC can perform to find the best position to ``align'' two time series. We set $max\_shift$ to $m$ to allow $k$-SC to find the optimal alignment across all possible shifts for each time series considered.

\begin{table}[h]
    \centering
    \makebox[\textwidth][c]{%
    \begin{tabular}{|l|l|l|l|l|l|l|}
    \hline
    Approach                & Metric    & Parameters & Default \\\hline
    SBD  No        &  - & - \\
    $k$-SC distance      & Yes        &   $max\_shift \in [0,\dots,m]$ & $max\_shift = m$ \\
    DTW    & No        &    $w \in [0,\dots,1]$ & $w = 1.0$\\
    Euclidean distance    & Yes        &  - & -\\
    soft-DTW    & No        &    $\gamma \in [0,\dots,\infty]$ & $\gamma = 1.0$\\
    \hline
    \end{tabular}
    }
    \vspace{0.25cm}
    \caption{Baseline Lloyd's-based models distance parameters.}
    \label{tab:lloyds-base-model-distance-configuration}
\end{table}
soft-DTW takes a parameter $\gamma$ which controls the smoothness of the gradient. $\gamma$ is challenging to set because small changes significantly impact results.~\cite{cuturi2017softdtw} experimented with four values of $\gamma$: $\{1.0, 0.1, 0.01, 0.001\}$. They found that smaller values of $\gamma$ often lead to barycentres getting stuck in bad local minima. However, their results also demonstrated that for some datasets, better results could be obtained with lower values of $\gamma$ (i.e., $0.01$ and $0.001$). Ultimately, they concluded, however, that in the average case, it was better to use a higher value of $1.0$, as it consistently converged to ``reasonable'' solutions. We, therefore, opt to use a value of $1.0$ to optimise for the average case.

DTW can use a window parameter $w$. For now, we set $w = 1.0$, meaning a full window will be used. 
Finally, the defined averaging techniques can also take additional parameters. For example, $k$-means-DBA uses DTW, which can be parameterised with a window. Similarly, $k$-SC averaging takes a $max\_shift$ parameter and $k$-means-soft-dba averaging takes a $\gamma$ parameter. Throughout all of our experiments in this thesis, unless explicitly stated otherwise, the same parameters specified for the distance computation are also applied in the averaging computations.

With the configuration and model definitions for our baseline experiment now established, we proceed with the experimentation and analysis of results. Results have been divided into two categories: combined test-train split and test-train split. 

\subsection{Combined test-train results}
\label{sec:lloyds-baseline-combined-test-train}

\begin{figure}[htb]
    \centering
            \begin{tabular}{c c}
    \includegraphics[width=0.5\linewidth]{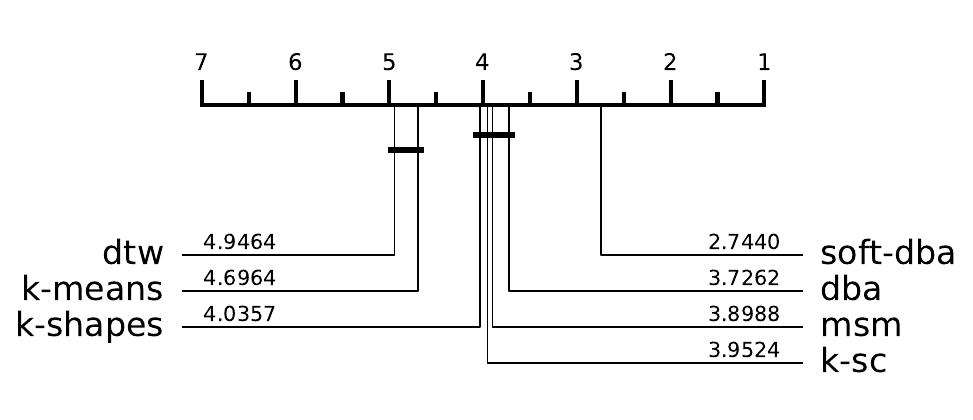} &
    \includegraphics[width=0.5\linewidth]{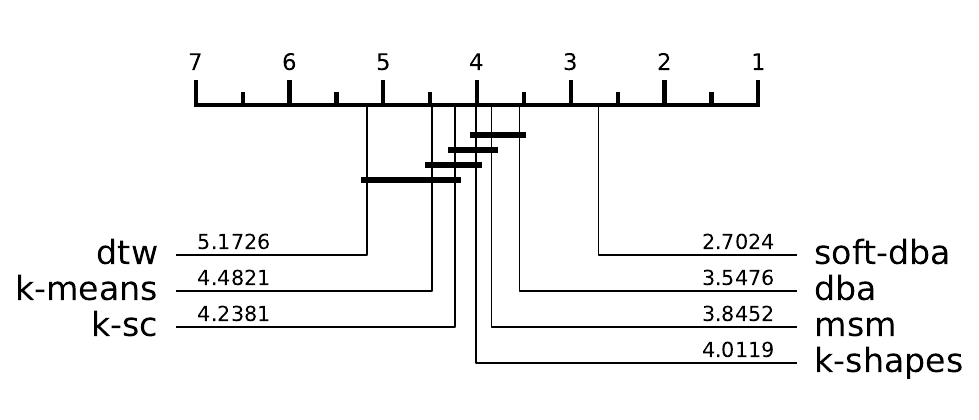}\\
    Accuracy & Average Rand Index \\
    \includegraphics[width=0.5\linewidth]{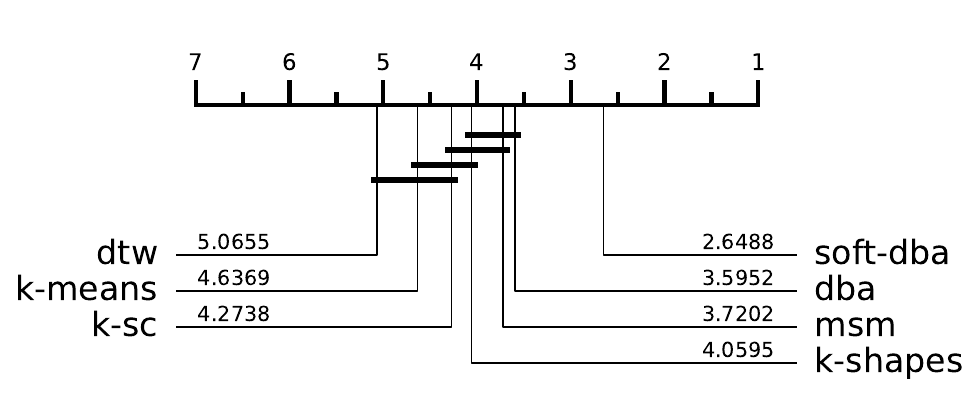} & 
    \includegraphics[width=0.5\linewidth]{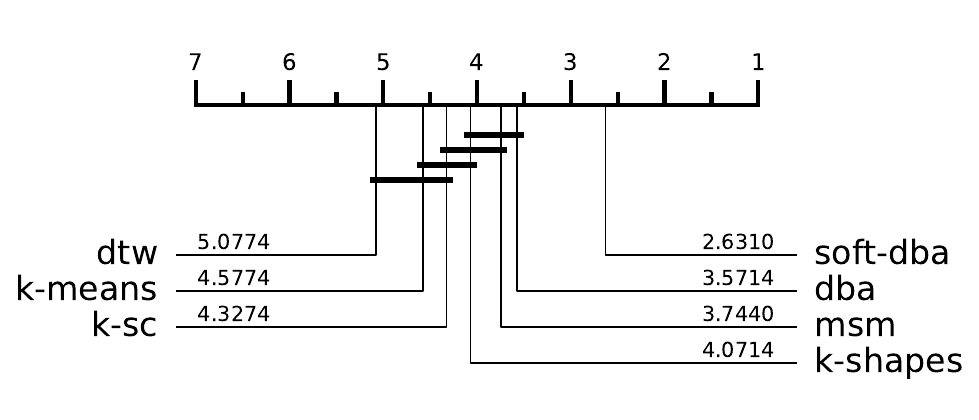} \\
    Average Mutual Information & Normalised Mutual Information \\
    
    \end{tabular}
   
    \caption{CD diagrams of Lloyd's-based algorithm over 84 datasets from the UCR archive using the combined test-train split. Datasets were excluded because $k$-means-soft-dba did not finish within our seven day runtime limit.}
    \label{fig:combined1}
\end{figure}

Figure~\ref{fig:combined1} presents the critical difference diagrams for seven clusterers using four clustering metrics for the combined test-train split. The average of each metric is given in Table~\ref{tab:combined1}. We observe that $k$-means-soft-dba significantly outperforms all other baseline Lloyd's clusterers across all evaluation metrics. However, soft-DBA only finished on 84 datasets within our maximum seven-day limit and took on average over 17 hours per dataset, and order of magnitude slower than any of the others. We observe that dtw is the worst performing clusterer. This is a confirmation of results presented in~\cite{holder24clustering}. There is no significant difference between k-shapes, msm and dba for any metric. Of these three, k-shapes is the fastest by an order of magnitude. 

\begin{table}[h]
\vspace{0.25cm}
\centering
\begin{tabular}{|l|l|l|l|l|l|l|}
\hline
 & ARI & AMI & CLAcc & NMI & RI & Average fit time (hrs)\\
\hline
dba & 0.264 & 0.307 & 0.593 & 0.318 & 0.694 & 1.62 \\
dtw & 0.186 & 0.228 & 0.532 & 0.241 & 0.639 & 1.39 \\
k-means & 0.202 & 0.247 & 0.538 & 0.260 & 0.672 &  0.0006\\
k-sc & 0.214 & 0.248 & 0.557 & 0.260 & 0.634 & 0.65 \\
k-shapes & 0.237 & 0.288 & 0.575 & 0.299 & 0.684 & 0.02 \\
msm & 0.229 & 0.277 & 0.568 & 0.288 & 0.679 & 0.29 \\
soft-dba & \textbf{0.303} & \textbf{0.342} & \textbf{0.624} & \textbf{0.352} & \textbf{0.710} & \textbf{17.62} \\
\hline
\end{tabular}
\caption{Clustering performance summary of seven algorithms over 84 UCR datasets using the combined test-train split. Fit time is total over }
\label{tab:combined1}
\end{table}

\begin{figure}[htb]
    \centering
            \begin{tabular}{c c}
    \includegraphics[width=0.5\linewidth]{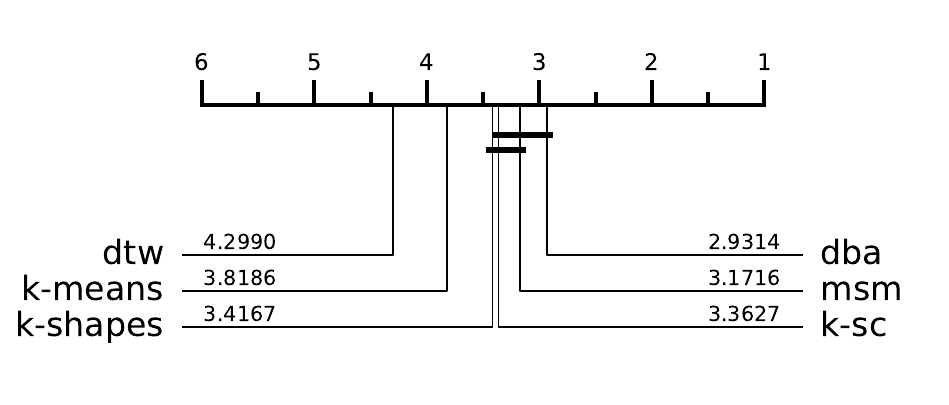} &
    \includegraphics[width=0.5\linewidth]{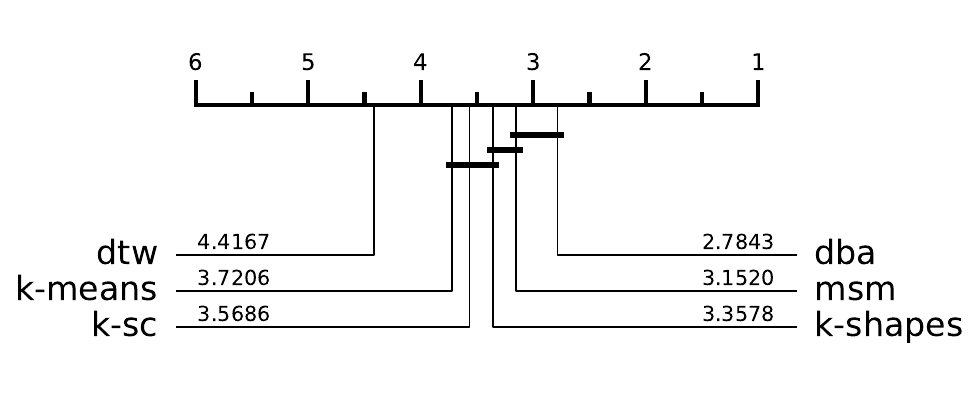}\\
    Accuracy & Average Rand Index \\
    \includegraphics[width=0.5\linewidth]{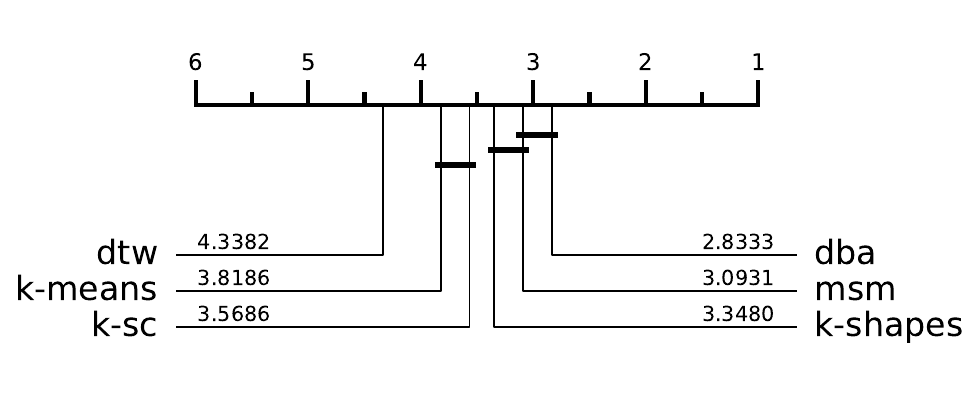} & 
    \includegraphics[width=0.5\linewidth]{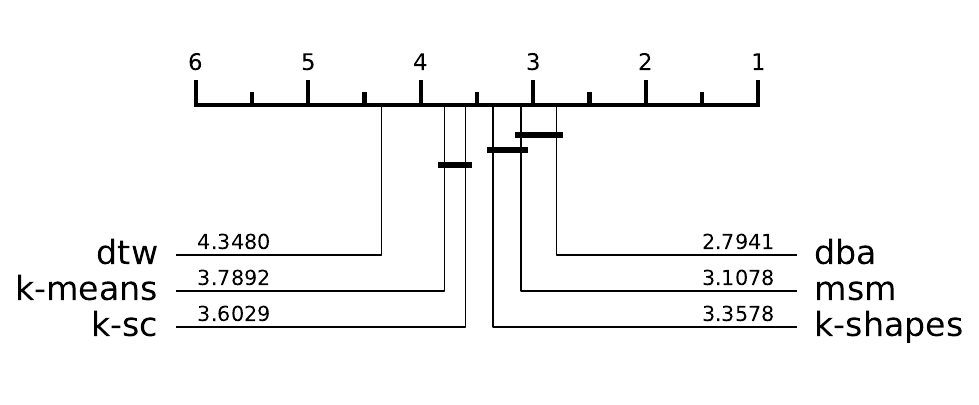} \\
    Average Mutual Information & Normalised Mutual Information \\    
    \end{tabular}
   
    \caption{Clustering performance summary of six algorithms over 102 UCR datasets using the combined test-train split.}
    \label{fig:combined2}
\end{figure}

We exclude soft-DBA and repeat the analysis to look at performance over more datasets. Figure~\ref{fig:combined2} presents the critical difference diagrams for our Lloyd's baseline experiment. We include results from $102$ of the $112$ datasets. $10$ datasets are excluded because dba exceeded our computational runtime limit. dba and msm form a top clique in three of the four metrics, with k-shapes and k-sc joining the top clique for accuracy. Both dtw and msm use the arithmetic mean to average. The difference in run time and performance between these clusterers demonstrates how inappropriate arithmetic averaging is for dtw. The run time of the distances is equivalent, so the differences are explained by dtw not converging: the dtw and averaging steps effectively work against each other and hindering the optimisation process. 

Overall, this baseline highlights the challenges of TSCL and helps set realistic expectations. In contrast, the current state-of-the-art model in TSC, HIVE-COTE 2.0~\citep{middlehurst21hc2}, achieves an average accuracy of $89.14\%$~\citep{middlehurst2024bakeoffredux} across 112 datasets from the UCR archive, compared to the $1NN$-Euclidean baseline, which achieves $68.62\%$ accuracy (\cite{dau19ucr}). This results in a classification accuracy difference of $20.52\%$. From our baseline experiment, it is evident that TSCL has yet to achieve this level of improvement over traditional approaches. Therefore, it is crucial to contextualise TSCL results in comparison to related fields such as TSC.

\subsection{Test-train split results}
\begin{figure}[htb]
    \centering
            \begin{tabular}{c c}
    \includegraphics[width=0.5\linewidth]{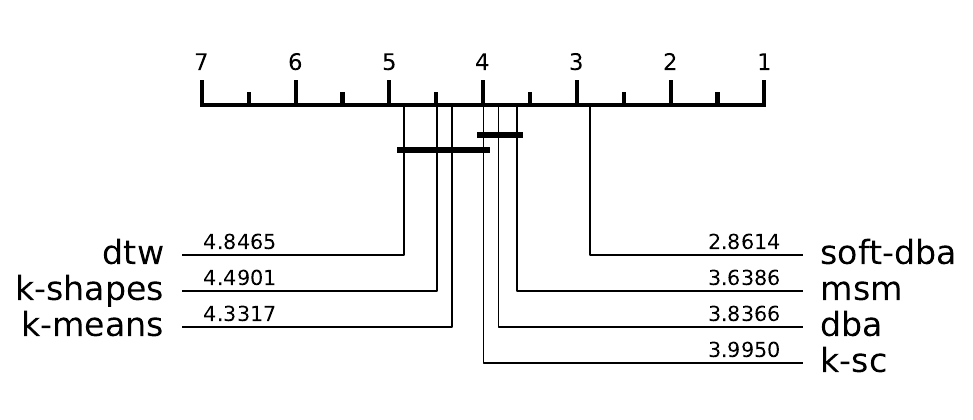} &
    \includegraphics[width=0.5\linewidth]{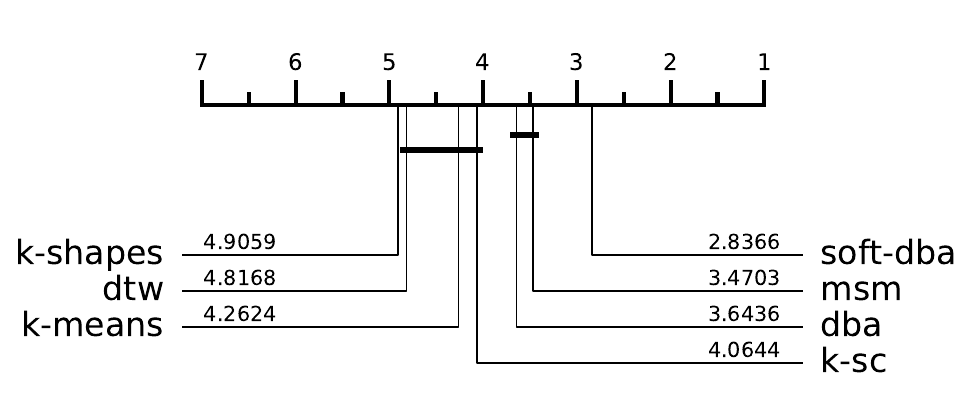}\\
    Accuracy & Average Rand Index \\
    \includegraphics[width=0.5\linewidth]{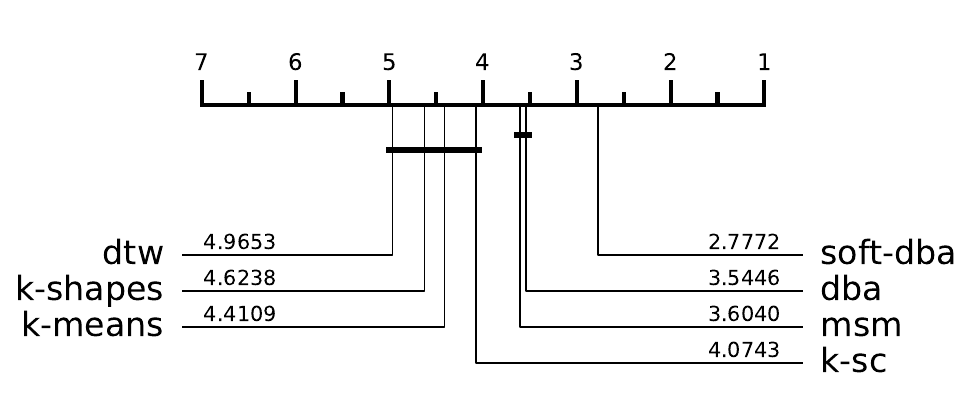} & 
    \includegraphics[width=0.5\linewidth]{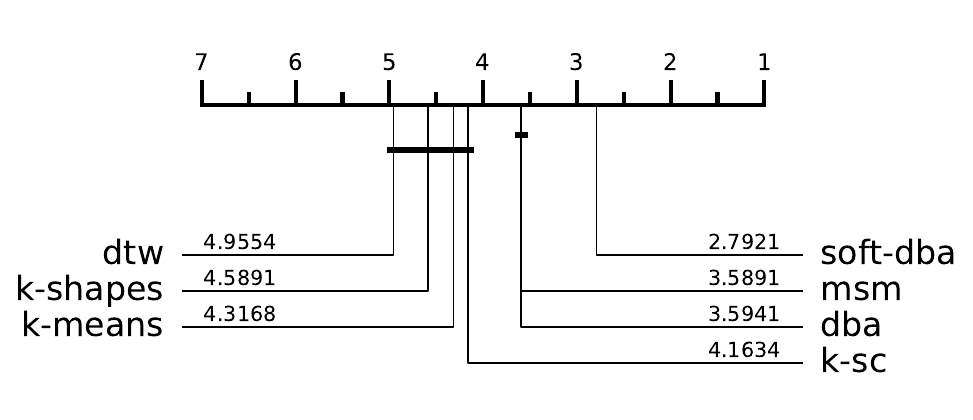} \\
    Average Mutual Information & Normalised Mutual Information \\
    \end{tabular}
   
    \caption{CD diagrams of Lloyd's-based algorithm over 75 datasets from the UCR archive using the combined test-train split. Datasets were excluded because $k$-means-soft-dba did not finish within our seven day runtime limit.}
    \label{fig:traintest1}
\end{figure}

We are also interested in performance when a model is trained and evaluated on separate data. Figure~\ref{fig:traintest1} shows the critical difference diagrams for four clustering metrics across 104 UCR datasets. As with the combined test-train split critical diagrams in Figure~\ref{fig:traintest1}, $k$-means-soft-dba significantly outperforms the other clusterers.

\begin{figure}[htb]
    \centering
            \begin{tabular}{c c}
    \includegraphics[width=0.5\linewidth]{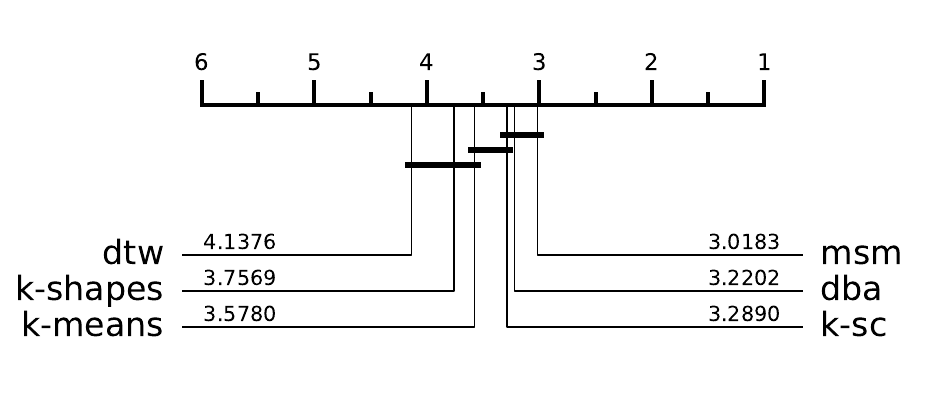} &
    \includegraphics[width=0.5\linewidth]{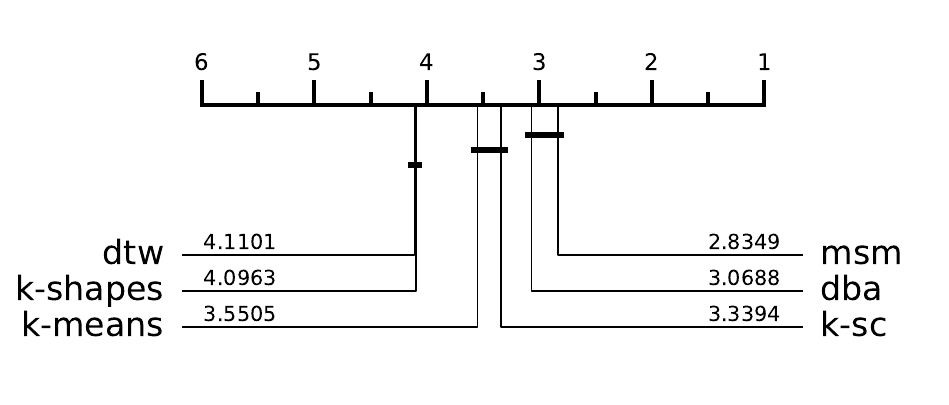}\\
    Accuracy & Average Rand Index \\
    \includegraphics[width=0.5\linewidth]{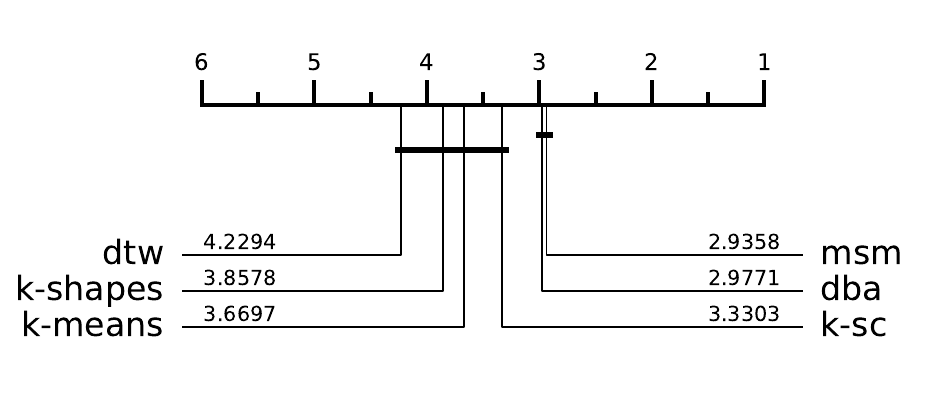} & 
    \includegraphics[width=0.5\linewidth]{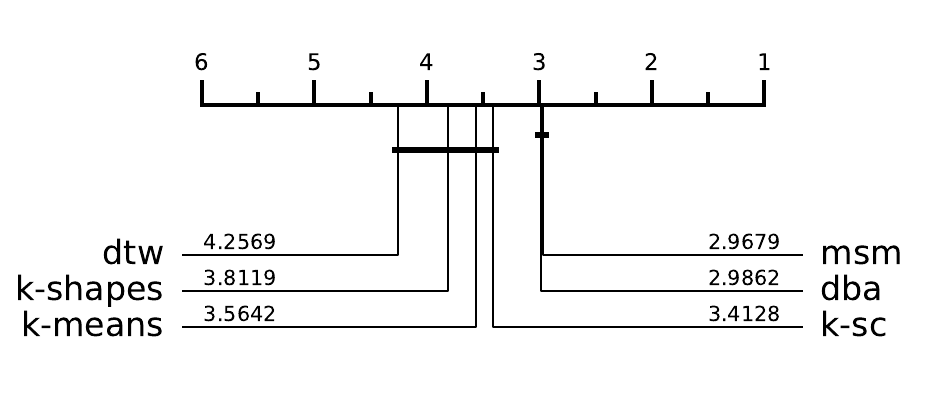} \\
    Average Mutual Information & Normalised Mutual Information \\    
    \end{tabular}
   
    \caption{Clustering performance summary of six algorithms over 102 UCR datasets using the combined test-train split.}
    \label{fig:traintest2}
\end{figure}

To maintain consistency with our combined test-train split evaluation, we now exclude $k$-means-soft-dba from our baseline analysis. Figure~\ref{fig:traintest2} presents the critical difference diagrams for our baseline experiments across 112 UCR archive datasets using the test-train split. The figure shows that $k$-means-dba remains the best-performing clusterer, though by a smaller margin. However, for CLACC, $k$-means-dba is not significantly different from $k$-sc and $k$-means-euclidean.

Another notable observation from Figure~\ref{fig:traintest2} is that $k$-shapes performs particularly poorly, with the worst average rank across all four clustering metrics. This is surprising, as $k$-shapes performed well on average in the combined test-train split experiments. This discrepancy potentially highlights a weakness of $k$-shapes: it may struggle to learn robust general representations of the data, leading to poorer performance on new, unseen data.

\begin{table}[h]
\vspace{0.25cm}
\centering
\begin{tabular}{|l|l|l|l|l|l|}
\hline
 & ARI & AMI & CLAcc & NMI & RI \\
\hline
dba & 0.234 & 0.286 & 0.563 & 0.308 & 0.693 \\
dtw & 0.175 & 0.228 & 0.522 & 0.251 & 0.645 \\
k-means & 0.184 & 0.232 & 0.525 & 0.256 & 0.673 \\
k-sc & 0.187 & 0.234 & 0.534 & 0.257 & 0.657 \\
k-shapes & 0.119 & 0.182 & 0.489 & 0.206 & 0.595 \\
msm & 0.219 & 0.270 & 0.553 & 0.291 & 0.682 \\
soft-dba & \textbf{0.254} & \textbf{0.311} & \textbf{0.588} & \textbf{0.331} & \textbf{0.699} \\
\hline
\end{tabular}
\caption{Lloyd's baseline experiment average ARI score on problems split by problem domain over $112$ datasets from the UCR archive using the test-train split.}
\label{tab:baseline-lloyds-test-train-ari-problem-domain}
\end{table}

\section{Conclusions}
\label{sec:conc}

Lloyd's algorithm is a popular approach for clustering, including clustering time series. Our review of the TSCL literature has shown that the community has adopted a range of experimental configurations, some of which are known to affect performance and are not always clearly specified. We have demonstrated the importance of initialisation and termination stages and specified an end-to-end distance based TSCL $k$-means algorithm where the same distance is consistently used in all steps. We have compared seven popular variants of TSCL $k$-means and identified that soft-dba gives the best clustering, but is bar far the slowest algorithm. We have verified the results presented in~\cite{holder24clustering} regarding the danger of using DTW naively with $k$-means: it performs worse than Euclidean distance when used with arithmetic averaging to find centroids. $k$-shapes is fast and performs well on the combined train-test data. However, it does much worse when using a train-test split. DBA is in the top clique without soft-dba, but takes significantly longer to run than MSM.

These findings are of interest in their own right, but also indicative of how a structured experimental regime can help standardise comparisons to find genuine areas of difference. Our implementation of Lloyd's clustering is available in the open soruce \texttt{aeon} toolkit, in addition to a wide range of alternative partitional and deep learning clustering algorithms. All experiments can be easily reproduced using the \texttt{tsml-eval}\footnote{https://github.com/time-series-machine-learning/tsml-eval} package, using the notebook as a guide. 

\bibliography{References/thesis} 

\end{document}